\documentclass{article}

\newcommand{\RR}{\mathbb{R}} 
\newcommand{\Let}[2]{\State #1 $\gets$ #2}

\usepackage{spconf,amsmath,graphicx}

\graphicspath{{plots/}}
\usepackage{subcaption}
\usepackage{wrapfig}
\usepackage{algorithm}
\usepackage[noend]{algpseudocode}

\usepackage{hyperref}       
\usepackage{url}            
\usepackage{booktabs}       
\usepackage{amsfonts}       
\usepackage{nicefrac}       
\usepackage{microtype, xcolor}      

\title{Contrastive Multivariate Singular Spectrum Analysis}
\name{Abdi-Hakin Dirie \qquad Abubakar Abid$^{\star}$ \qquad James Zou$^{\star}$}
\address{$^{\star}$ Stanford University, Department of Electrical Engineering}

\begin{document}
%
\maketitle
\begin{abstract}
We introduce Contrastive Multivariate Singular Spectrum Analysis, a novel unsupervised method for dimensionality reduction and signal decomposition of time series data. By utilizing an appropriate background dataset, the method transforms a target time series dataset in a way that evinces the sub-signals that are \textit{enhanced} in the target dataset, as opposed to only those that account for the greatest variance. This shifts the goal from finding signals that \emph{explain} the most variance to signals that \emph{matter} the most to the analyst. We demonstrate our method on an illustrative synthetic example, as well as show the utility of our method in the downstream clustering of electrocardiogram signals from the public MHEALTH dataset.
\end{abstract}
\section{Introduction}
\label{sec:intro}

Unsupervised dimensionality reduction is a key step in many applications, including visualization \cite{maaten2008visualizing} \cite{mcinnes2018umap}, clustering \cite{cohen2015dimensionality} \cite{niu2011dimensionality}, and preprocessing for downstream supervised learning \cite{pechenizkiy2004pca}. Principal Component Analysis (PCA) is one well-known technique for dimensionality reduction, which notably makes no assumptions about the ordering of the samples in the data matrix $X \in \RR^{N \times D}$. Multivariate Singular Spectrum Analysis (MSSA) \cite{hassani2013multivariate} is an extension of PCA for time series data, which been successfully applied in applications like signal decomposition and forecasting \cite{hassani2009forecasting} \cite{mahmoudvand2015forecasting} \cite{patterson2011multivariate}. In MSSA, each row is read at a certain time step, and thus is influenced by the ordering of the samples. MSSA works primarily by identifying key oscillatory modes in a signal, which also makes it useful as a general-purpose signal denoiser. However, MSSA (like PCA, upon which it is based) is limited to finding the principal components that capture the maximal variance in the data. In situations where the information of interest explains little overall variance, these methods fail to reveal it. Recently, extensions like contrastive PCA (cPCA) \cite{abid2018exploring, zou2013contrastive,ge2016rich} have shown that utilizing a background dataset $Y \in \RR^{M \times D}$ can help better discover structure in the foreground (target) $X$ that is of interest to the analyst.

\begin{figure}[htb]
    \centering
    \includegraphics[width=\columnwidth]{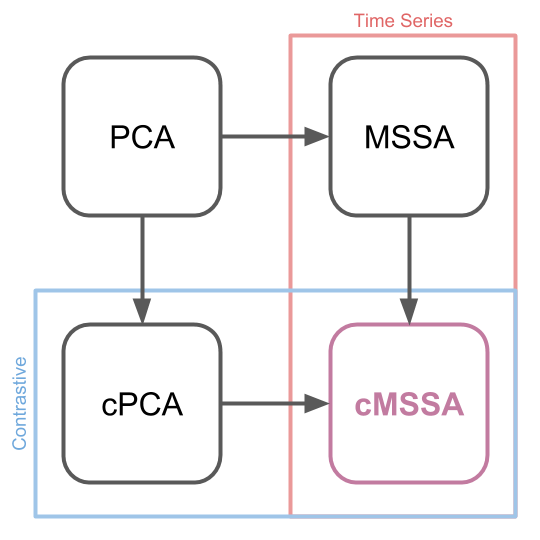}
    \caption{Schematic illustrating the relations among PCA, cPCA, MSSA, and cMSSA.}
    \label{fig:diagram}
\end{figure}

Contrastive Multivariate Singular Spectrum Analysis (cMSSA) generalizes cPCA and applies it to time series data. Figure~\ref{fig:diagram} visualizes the relationships between the four methods. As a contrastive method, cMSSA emphasizes salient and unique sub-signals in time series data rather than just the sub-signals comprise the majority of the structure. So while standard MSSA is useful for denoising a signal, cMSSA additionally ``denoises'' signals of structured but irrelevant information.

\section{Contrastive Multivariate Singular Spectrum Analysis}\label{cmssa}

\textbf{Standard MSSA}
Consider a centered one-channel times series $\mathbf{x} \in \RR^T$. We construct a Hankel matrix $H_\mathbf{x} \in \RR^{T' \times W}$ with window size $W$ as follows:
\[
H_\mathbf{x} = 
\begin{pmatrix}
x_1 & x_2 & \ldots & x_W \\
x_2 & x_3 & \ldots & x_{W+1} \\
\vdots & \vdots & \ddots & \vdots \\
x_{T'} & x_{T'+1} & \ldots & x_T \\
\end{pmatrix}
\]
where $T' = T-W+1$.
To extend to the multivariate case, let $X \in \RR^{T \times D}$ be a $D$-channel time series that runs for $T$ steps. We construct the Hankelized matrix $H_X$ with window $W$ by horizontally concatenating the per-channel Hankel matrices into a $T'$-by-$DW$ matrix:
$H_X = [H_{\mathbf{x}^{(1)}} ; H_{\mathbf{x}^{(2)}} ; \ldots ; H_{\mathbf{x}^{(D)}}]$. Next we compute the covariance matrix $C_X \in \RR^{DW\times DW}$ for $H_X$. The next step is to perform the eigendecomposition on $C_X$, yielding $DW$ eigenvectors. Of these we take the top $K$ vectors with the largest corresponding eigenvalues. We denote $\mathbf{e}^{(k)}$ as the eigenvector with the $k$th largest eigenvalue. We collect the vectors into a matrix $E \in \RR^{DW \times K}$.

To transform our original time series $X$, we have two options: (a) Project $X$ into the principal component (PC) space defined by $E$:
$A = H_X E$ or (b) use $A$ to compute the $k$th reconstructed component (RC) $R^{(k)}$ as done in the SSA literature:

\[
R^{(k)}_{tj} = \frac{1}{W_t} \sum^{U_t}_{t' = L_t} A_{t-t'+1, k} \cdot \mathbf{e}^{(k)}_{(j-1)W + t'}
\]

where $L_t = \max(1, t-T+W)$, $U_t = \min(t, W)$, and $W_t = U_t - L_t + 1$. The rows of $R^{(k)}$ are indexed by time $t \in \{1,\ldots,T\}$ and the columns by channel $j \in \{1,\ldots,D\}$. Summing up the reconstructed components reproduces a denoised version of the original signal. For our purposes, we opt instead to take the horizontal concatenation of the reconstructed components as the second transform:
$R = [R^{(1)} ; R^{(2)} ; \ldots ; R^{(K)}].$
To handle multiple time series, one simply vertically stacks each Hankelized matrix. The algorithm proceeds identically from there.

\bigskip
\noindent \textbf{Contrastive MSSA}
The modification to MSSA we introduce is via a new variable $\alpha \geq 0$ we call the \emph{contrastive} hyperparameter. We construct $H_Y$ for another $D$-channel times series $Y$ (the background data) via the same process. It is not required that $X$ and $Y$ run for the same number of time steps, only that their channels are aligned. We compute a contrastive covariance matrix $C = C_X - \alpha C_Y$ and perform the eigendecomposition on $C$ instead of $C_X$. The intuition for this is that by subtracting out a portion of the variance in $Y$, the remaining variance in $X$ is likely to be highly specific to $X$ but not $Y$.
This is the key additional mechanism behind cMSSA --- if $\alpha = 0$, then no contrast is performed, and cMSSA reduces down to just MSSA.

\begin{algorithm}[htb]
\caption{Spectral $\alpha$-Search
}
\label{algo:gen_alpha}
\begin{algorithmic}[1]

\Require Minimum $\alpha$ to consider $\alpha_{\min}$, maximum $\alpha$ to consider $\alpha_{\max}$, number of $\alpha$s to consider $n$, number of $\alpha$s to return $m$, foreground signal $X$, background signal $Y$, window $W$, and number of components $K$.

\Procedure{}{}
    \Let{$Q$}{\textsc{LogSpace}($\alpha_{\min}$, $\alpha_{\max}$, $n$) $\cup \{0\}$}
    \For{$\alpha^{(i)} \in Q$}
        \Let{$H_X$, $H_Y$}{\textsc{Hankel}($X$, $W$), \textsc{Hankel}($Y$,$W$)}
        \Let{$C_X$, $C_Y$}{\textsc{Cov}($H_X$), \textsc{Cov}($H_Y$)}
        \Let{$E^{(i)}$}{\textsc{EigenDecomp}($C_X - \alpha^{(i)}C_Y$, $K$)}
    \EndFor
    \Let{$S$}{\textsc{Empty}($\RR^{n+1 \times n+1}$)}
    \For{$i \in \{1, \ldots, n+1\}$, $j \in \{i, \ldots, n+1\}$}
        \Let{$S_{i,j}, S_{j,i}$}{$\left\lVert {E^{(i)}}^T E^{(j)} \right\rVert_*$}
    \EndFor
    \Let{$Z$}{\textsc{SpectralCluster}($S$, $Q$, $m$)}
    \Let{$Q^*$}{\{0\}}
    \For{$z \in Z$}
        \If{$0 \notin z$}
            \Let{$\alpha^*$}{\textsc{ClusterMediod}($z$, $S$)}
            \Let{$Q^*$}{$Q^* \cup \{\alpha^*\}$}
        \EndIf
    \EndFor
    \\
    \Return{$Q^*$, set of $m$ best $\alpha$s, including zero.}
\EndProcedure
\end{algorithmic}
\end{algorithm}

The choice of $\alpha$ is non-trivial.  Algorithm~\ref{algo:gen_alpha} outlines a routine for auto-selecting a small number of promising values for $\alpha$. Because cMSSA is designed to assist data exploration, Algorithm~\ref{algo:gen_alpha} uses spectral clustering to identify a diverse set of $\alpha$ values corresponding to diverse eigenspaces and representations of  $X$. The procedure works by first generating a large number of $\alpha$s spread evenly in log-space. For each candidate $\alpha$, we use cMSSA to compute its corresponding eigenvector matrix $E$. The procedure then performs spectral clustering, which requires a pairwise distance matrix as input. The distance metric used takes the nuclear norm of the matrix computed by multiplying the eigenvector matrices $E$ for any pair of $\alpha$s. After specifying the number of clusters desired, we take the mediod $\alpha$ of each cluster and return them as output. We always include 0 in this set, as the analyst may want to perform their analysis without contrast as control.

\section{Experiments}\label{experiements}

\begin{figure}[htb]
\centering
\begin{subfigure}{\columnwidth}
    \centering
    \includegraphics[width=\columnwidth]{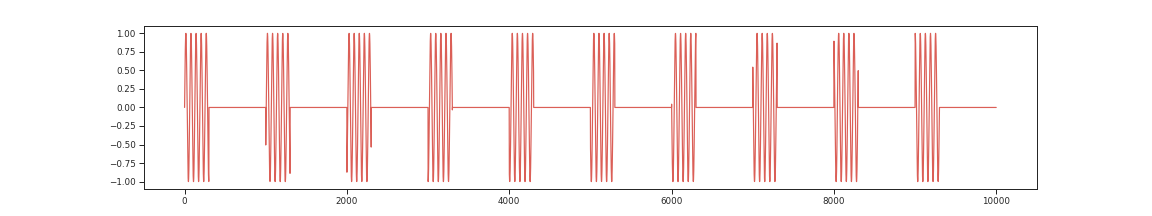}
    \caption{Sub-signal specific to the foreground data $X$, which is of much lower amplitude than the other sinusoidal sub-signals in $X$.}
    \label{fig:syn_sub}
\end{subfigure}
\begin{subfigure}{\columnwidth}
    \centering
    \includegraphics[width=\columnwidth]{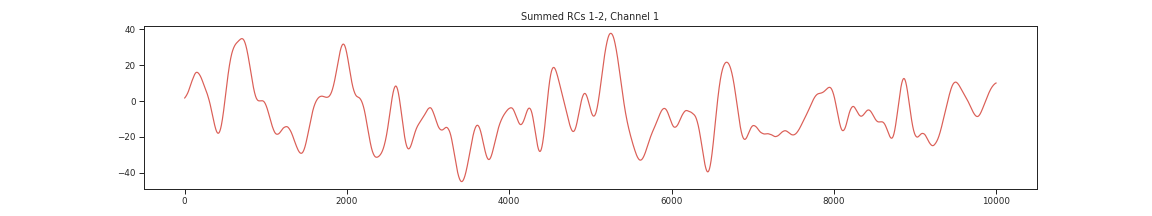}
    \caption{Without contrast, the reconstructed time series consists of the high-amplitude sinusoidal sub-signals in $X$.}
    \label{fig:syn_x_rcs}
\end{subfigure}
\begin{subfigure}{\columnwidth}
    \centering
    \includegraphics[width=\columnwidth]{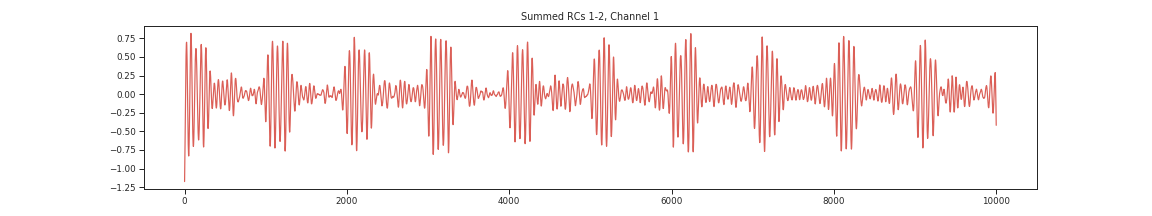}
    \caption{With contrast, the reconstructed time series is able to identify the unique sub-signal in $X$.}
    \label{fig:syn_x_rcs_contrast}
\end{subfigure}

\caption{Results of a synthetic experiment that demonstrates that cMSSA is able to identify unique sub-signals in a time series, even when they are of much lower amplitude than background components.}

\label{fig:syn_exp}
\end{figure}

\textbf{Synthetic example}
To illustrate cMSSA, we present a simple synthetic example. We generate an artificial one-channel signal $Y$ by sampling 500 sinusoids with different frequencies, amplitudes, phases, and vertical shifts. White Gaussian noise sample from $\mathcal{N}(0,1)$ is added in as well. We generate $X$ in the same manner, but add in a very specific sub-signal (Figure~\ref{fig:syn_sub}) that has comparatively low variance compared to the whole time series. The signals $X$ and $Y$ are generated independently as to rule out simple signal differencing as an explanation. We take $X$ as foreground and $Y$ as background.

We set $W=100$, $\alpha=2$, and use only the top $K=2$ RCs. Fig.~\ref{fig:syn_exp} displays the reconstructions computed by MSSA versus cMSSA, alongside the sub-signal that was injected into $X$. Specifically, we see that the cMSSA reconstruction shown in Fig.~\ref{fig:syn_x_rcs_contrast} yields a noisy approximation of the sub-signal of interest,  Fig.~\ref{fig:syn_sub}. The variance of the noise here is comparable to the variance of the sub-signal---more noise would eventually overpower cMSSA's ability to extract the sub-signal.

\bigskip
\noindent \textbf{Clustering of electrocardiograms}
The data used in this experiment is taken from the public MHEALTH dataset \cite{banos2014mhealthdroid}. In the dataset, 10 individuals were asked to perform 12 physical activities as several sensors recorded varied motion data. The researchers also collected two-lead electrocardiogram (ECG) readings, which we take as dual-channel time series data. In addition to the 12 activities, there is a 13th \textsc{NULL} class that represents ECG signals collected between each activity but which don't have labels themselves. To increase the number of individual time series, we partition each one in half. 

    
    

For our experiments, the foreground data are all time series labelled as either \textsc{Jogging}, \textsc{Running}, \textsc{Jumping}, or \textsc{Cycling}, 20 time series each for a total of 80. These four, being the more cardio-intensive of the 12, had much more signal activity that would be needed to be sifted through, exactly the type of environment cMSSA is intended to handle. For background data, we take all 272 time series belonging to the \textsc{NULL} class.

To evaluate the effectiveness of cMSSA over its non-contrastive counterpart, we run both cMSSA and MSSA with a variety of hyperparameter settings. For each fitted model, we transform the foreground data to both the PC and RC spaces. Once the transformations are had, we perform  spectral clustering into 4 clusters and compare the resulting clusters to the activity labels on the time series data, which were hitherto withheld from the algorithms. There are 3 hyperparameters: the window size $W \in \{8, 16, 32, 84, 18\}$, the number of desired components $K \in \{1,2,4,6,8,10,12,14,16,18,20\}$, and the contrastive parameter $\alpha$. We set $K$ only if the value is less than or equal to $DW$ (where $D=2$ in this case). For $\alpha$, we used our automatic routine to compute five key values to try for each setting of $W$ and $K$. For each run of the routine, a total of 300 candidate $\alpha$s we considered, with the minimum and maximum $\alpha$s being $10^{-3}$ and $10^{3}$, respectively. Of the five ultimately returned, one was zero, representing standard MSSA. Altogether, we run 530 experiments, 106 of which are standard MSSA, and the remaining cMSSA.

The spectral clustering requires an affinity matrix $S \in \RR^{N \times N}$ which contains the similarities between any pair of time series, where $N$ is the number of times series we wish to cluster. Let $X^{(i)}$ and $X^{(j)}$ be two time series. Using the FastDTW metric \cite{salvador2007toward} with a euclidean norm\footnote{FastDTW is not a symmetric metric, so we take the minimum between the two orderings of the operands.}, we define the similarity $S_{ij}$ to be
$
\frac{1}{\textsc{FastDTW}(X^{(i)}, X^{(j)}) + 1}.
$
The cluster evaluation uses the well-rounded BCubed metric \cite{amigo2009comparison} to compute the precision, recall, and F1 harmonic mean for a particular cluster prediction. We also perform the evaluation in the model-free sense where we simply cluster the time series with no transformation as a basic baseline.

\begin{table}[htb]
  \caption{Best cMSSA and MSSA results in terms of maximum F1 score. Model-free clustering baseline also included. For the best MSSA and cMSSA models (with $\alpha$ automatically selected via Algorithm 1), PC transform outperformed RC transform. Best result per metric (precision, recall, or F1) is bolded.}
  \label{tab:mhealth_best}
  \centering
    \begin{tabular}{l | c c c}
    \toprule
    Model &
    $W$ &
    $K$ &
    P / R / F1 \\
    \midrule
    
    Model-free & - & - & 50.49 / 48.82 / 49.54 \\
    MSSA & 16 & 16 & 57.67 / 64.63 / 60.95 \\
    cMSSA ($\alpha = 12.41$) & 128 & 1 & \textbf{65.44} / \textbf{75.88} / \textbf{70.27} \\
    
    \bottomrule
    \end{tabular}
\end{table}

Table~\ref{tab:mhealth_best} reports the best representative contrastive and non-contrastive models, comparing both to the model-free baseline. We observe a number of things. First, both MSSA and cMSSA outperform the model-free baseline. Second, cMSSA has 9-10 point gains over cMSSA in each of precision, recall, and F1. Third, both find that using $A$ over $R$ as the transform yielded better results.
Finally, of the $DW$ number of PCs available, MSSA gets its best performance using half (16 out of 32), while cMSSA only uses one PC out of the maximum of 256 available. This highlights an interesting efficiency of cMSSA. By filtering out unnecessary components, the remaining not only account for less signal variance, but provide diminishing returns with each subsequent component used.

\begin{figure}[htb]
    \centering
    \includegraphics[width=\columnwidth]{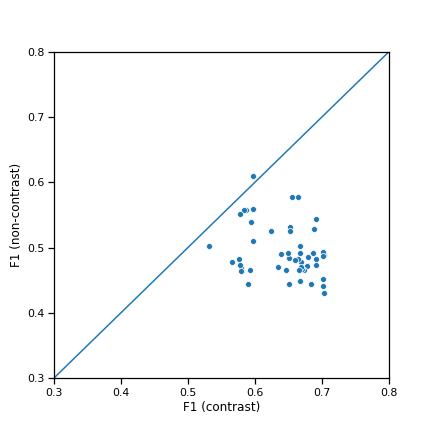}
    \caption{Plot of paired F1 scores. Each point is for a particular setting of $W$ and $K$. The contrastive F1 score used is the maximum of the four runs (one per automatically selected $\alpha$) for that setting of the hyperparams. $x=y$ line drawn as guidance. The points look at only those where the transform used is $A$.}
    \label{fig:f1_vs_F1_A}
\end{figure}

Figure~\ref{fig:f1_vs_F1_A} shows a more granular view of the general gains to be had from using cMSSA. For a particular setting of $W$ and $K$, we plot the F1 score for the non-contrastive case vs the contrastive case. Due to the four values of $\alpha$s used in the contrastive case, we take the model that had the greatest F1.  Points below the diagonal line mean that the contrast was useful for a particular setting of the hyperparameters.

\begin{figure}[htb]
\begin{subfigure}{0.46\columnwidth}
\begin{subfigure}{\columnwidth}
    \includegraphics[width=\columnwidth]{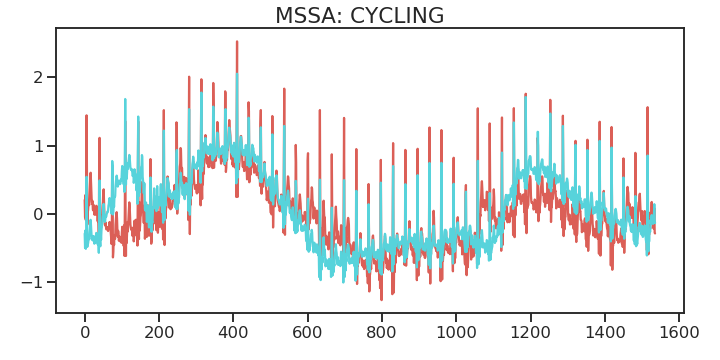}
    \label{fig:random_cycling_k16_no_contrast_best_compare}
\end{subfigure}
\begin{subfigure}{\columnwidth}
    \includegraphics[width=\columnwidth]{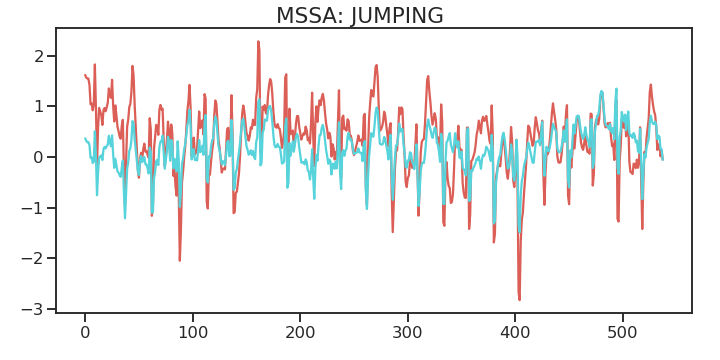}
    \label{fig:random_jumping_k16_no_contrast_best_compare}
\end{subfigure}
\begin{subfigure}{\columnwidth}
    \includegraphics[width=\columnwidth]{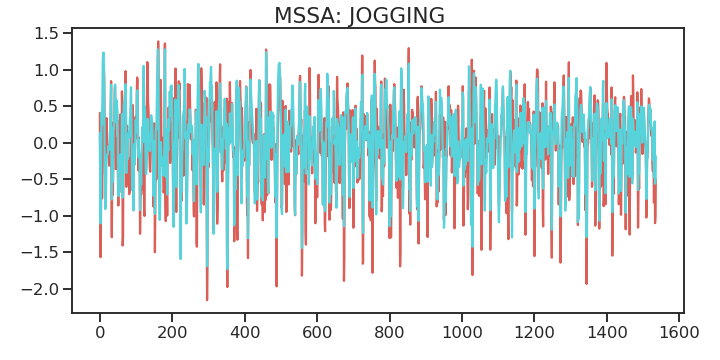}
    \label{fig:random_jogging_k16_no_contrast_best_compare}
\end{subfigure}
\begin{subfigure}{\columnwidth}
    \includegraphics[width=\columnwidth]{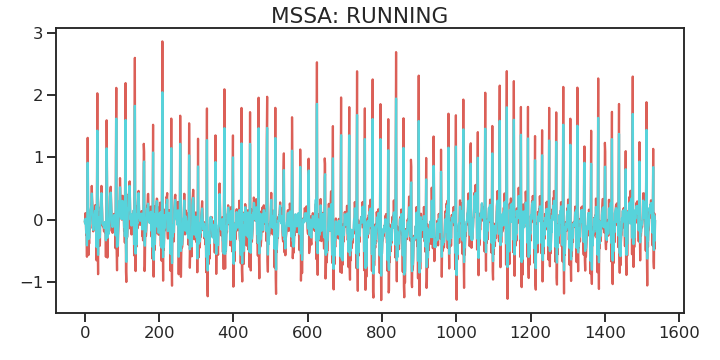}
    \label{fig:random_running_k16_no_contrast_best_compare}
\end{subfigure}
\end{subfigure}
\qquad
\begin{subfigure}{0.46\columnwidth}
\begin{subfigure}{\columnwidth}
    \includegraphics[width=\columnwidth]{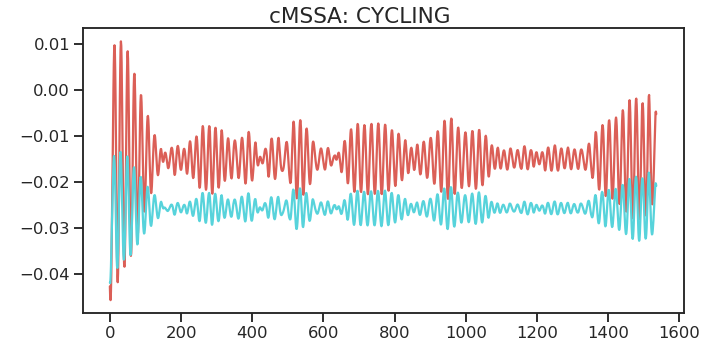}
    \label{fig:random_cycling_k1_contrast_best_compare}
\end{subfigure}
\begin{subfigure}{\columnwidth}
    \includegraphics[width=\columnwidth]{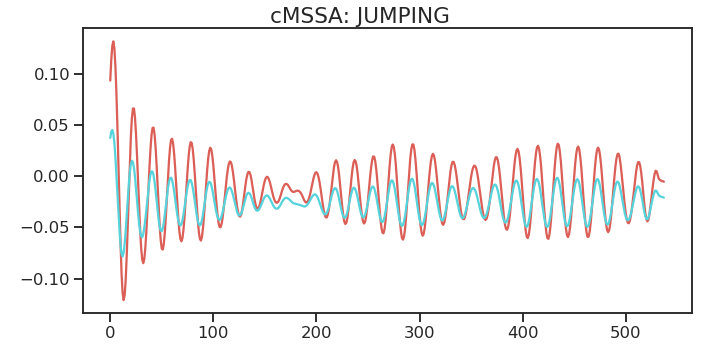}
    \label{fig:random_jumping_k1_contrast_best_compare}
\end{subfigure}
\begin{subfigure}{\columnwidth}
    \includegraphics[width=\columnwidth]{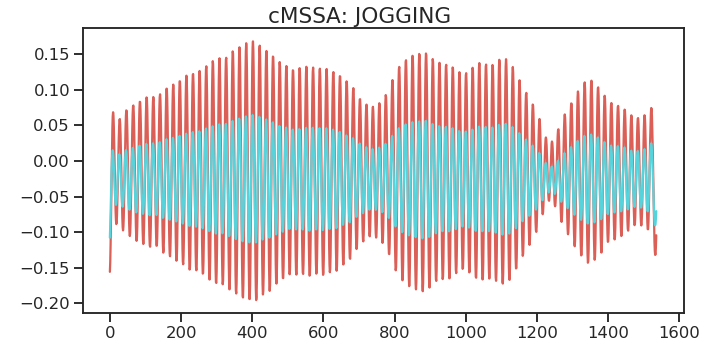}
    \label{fig:random_jogging_k1_contrast_best_compare}
\end{subfigure}
\begin{subfigure}{\columnwidth}
    \includegraphics[width=\columnwidth]{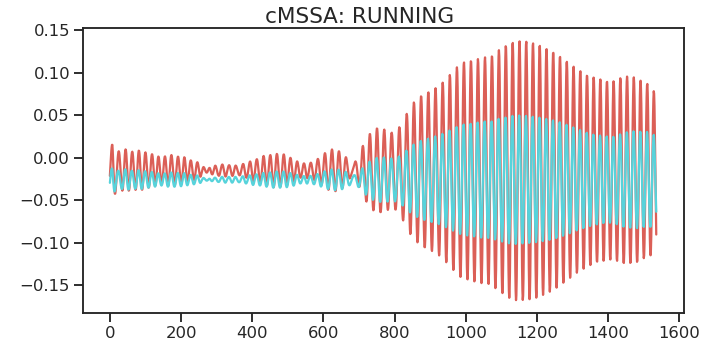}
    \label{fig:random_running_k1_contrast_best_compare}
\end{subfigure}
\end{subfigure}

\caption{Reconstructed time series after performing MSSA ($W=16$, $K=16$) and cMSSA ($W=128$, $K=1$, $\alpha = 12.41$). Each row is the same random signal reconstructed using MSSA (left) and cMSSA (right), one row per activity. The two colors correspond to the dual recording channels. 
}
\label{fig:random_compare}
\end{figure}

Finally, Figure~\ref{fig:random_compare} shows a visual comparison of MSSA versus cMSSA, using their respective hyperparameters settings as shown in Table~\ref{tab:mhealth_best}. Each row depicts how a random signal is processed with contrast on or off. We immediately see that cMSSA finds simpler signals than those found by MSSA. In the case of MSSA, the processed signals do not look substantially different from the originals. This is due to the fact that the high variance signals are shared across activities, so MSSA favors them during reconstruction. This is not the case with cMSSA, which identifies the differentiating signals that can disambiguate the activities.

\section{Conclusion}

We have developed cMSSA, a general tool for dimensionality reduction and signal decomposition of temporal data. By introducing a background dataset, we can efficiently identify sub-signals that are enhanced in one time series data relative to another. In an empirical experiment, we find that for virtually any setting of the hyperparameters, cMSSA is more effective at unsupervised clustering than MSSA, contingent on appropriate choices for the foreground and background data. It is worth emphasizing that cMSSA is an unsupervised learning technique. It does \emph{not} aim to discriminate between time series signals, but rather discover structure and sub-signals within a given time series more effectively by using a second time series as background. This distinguishes it from various discriminant analysis techniques for time series that are based on spectral analysis \cite{maharaj2014discriminant}, \cite{krafty2016discriminant}.

Some basic heuristics should be kept in mind when choosing to use cMSSA. First, the data ideally should exhibit periodic behavior, as MSSA (and by extension, cMSSA) is particularly well suited to finding oscillatory signals. Second, the data of interest $X$ and background $Y$ should not be identical, but should share common structured signal such that the contrast retains some information in the foreground. As an example, the ECG foreground data consisted of subjects performing very specific activities, whereas the background consisted of unlabelled ECG signals in which the participants performed no specific activity. We would expect a good amount of overlap in signal variance, but signals specific to the four activities would be under-represented in the background. Thus contrast is a plausible way to extract this signal.

Finally, we note that the only hyper parameter of cMSSA is the contrast
strength, $\alpha$. In our default algorithm, we developed an automatic
subroutine that selects
the most informative values of $\alpha$. The experiments
performed used the automatically
generated values. We believe that this default will be sufficient
in many use cases of cMSSA, but the user may also
set specific values for $\alpha$ if more granular exploration
is desired.




\vfill\pagebreak

\nocite{*}
\bibliographystyle{IEEEbib}

\end{document}